\pdfoutput=1

\documentclass[11pt]{article}

\usepackage[]{acl}

\usepackage{times}
\usepackage{latexsym}

\usepackage[T1]{fontenc}

\usepackage[utf8]{inputenc}

\usepackage{microtype}

\usepackage{amsmath}

\usepackage[show]{notes}
\newcommand{\ignore}[1]{}
\newcommand{\com}[1]{}

\usepackage{graphicx}   
\usepackage{tikz}       
\usepackage{subfig}
\usepackage{multirow}   
\usepackage{bchart}     

\usepackage[super]{nth}

\author{Eyal Arviv \and Oren Tsur\\
    Software and Information Systems Engineering \\
  Ben Gurion University of the Negev \\
  \texttt{eyalar@post.bgu.ac.il} ~~~ \texttt{orentsur@bgu.ac.il} \\
  }

\title{How to Do Things without Words: Modeling Semantic Drift of Emoji}

\begin{document}
\maketitle
\begin{abstract}
Emoji have become a significant part of our informal textual communication. Previous work, addressing the societal and linguistic functions of emoji, overlooked the relation between the semantics and the visual variations of the symbols. In this paper we model and analyze the semantic drift of emoji and discuss the features that may be contributing to the drift, some are unique to emoji and some are more general. Specifically, we explore the relations between graphical changes and semantic changes.

\end{abstract}

\paragraph{Warning:} Some readers may find some of the words and emoji interpretations offensive.

\section{Introduction}
\label{sec:intro}

Emoji have become a significant part of our informal textual communication. 
One of the major linguistic functions served by emoji\footnote{The term `emoji' refers to the set of codified pictograph (the `language'), to a specific icon (singular), and to a set of icons (plural).} is the re-introduction of physical-like gestures in textual messaging \cite{evans2017emoji,mcculloch2019because}. While Emoji is not a `natural language' in the typical sense, it is a linguistic phenomena that can be addressed from a (computational-) linguistic perspective. Moreover, the unique nature and inherent limitations of the use of emoji, provide a distinctive perspective for understanding various linguistic phenomena such as lexcicalization, gesturing, ambiguity and semantic drift. In this paper we focus on temporal distributional semantics of emoji. To the best of our knowledge, this is the first work to address semantic drift of emoji with respect to graphical variation across time and platform.

\vspace{-0.2cm} 
\paragraph{The Emoji Echo-system} The Unicode Consortium started adding Emoji to the Unicode standard at 2010, assigning a code to a short description\footnote{The complete list of codes and descriptions can be found at: \href{https://unicode.org/emoji/charts/full-emoji-list.html}{https://unicode.org/emoji/charts/full-emoji-list.html}.}. For example, \texttt{\footnotesize U+1F602} is described as \texttt{\footnotesize face with tears of joy}, \texttt{\footnotesize U+1F9D0} as \texttt{\footnotesize face with monocle}, and the \texttt{\footnotesize U+1F36A} code is simply defined as \texttt{\footnotesize cookie}. The graphic realization of the description is left for the discretion of the designers of each specific platform. For example, the graphic interpretation of the \texttt{\footnotesize cookie} code (\texttt{\footnotesize U+1F36A}) ranges from variations of single chocolate-chip cookie to a pair of crackers: \includegraphics[height=1em]{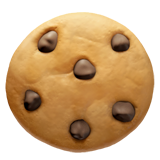}, \includegraphics[height=1em]{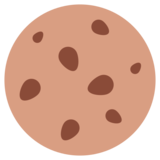}, and     \includegraphics[height=1em]{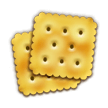} ( iOS 13.2, Twitter's Twemoji 12.2, and Samsung Touchwiz 7.1, respectively).

The exact meaning of an emoji depends on the social and textual contexts in which it is used -- the smiley emoji, for instance, can be used to show empathy, express amusement or to ease tension. The pistol emoji can be used to convey frustration or amusement (``you are killing me''), bragging (``I killed it'') or a proper threat (``I WILL kill you'') -- a disturbing illocutionary act 
 \cite{austin1975things,salgueiro2010promises}, performed without words. Consequently, the accurate interpretation of an emoji may have significant legal implications, possibly landing a defendant accused of murder a life in jail, if the emoji is considered a threat -- proving premeditation of the murder \cite{goldman2018emojis}. Similarly, if the semantics of the  \includegraphics[height=1em]{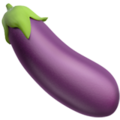} have drifted from a vegetable to a sexual reference (which is, indeed, the case), sending this emoji to a colleague could be perceived as a sexual misconduct.

\ignore{ \includegraphics[height=1em]{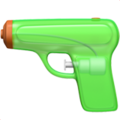},  \includegraphics[height=1em]{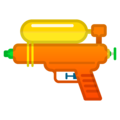}, \includegraphics[height=1em]{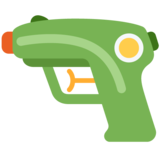} and \includegraphics[height=1em]{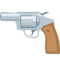} are the graphic interpretations of \footnotesize{\texttt{U+1F52B}} in iOS v13.2, Android v10, Twitter v12.2, and Facebook v2.0, respectively. }

\vspace{-0.2cm} 
\paragraph{Semantic Drift} Language keeps changing. Semantic drift, often referred to as semantic shift or semantic change, is the process in which words change their meaning over time \cite{hock2009language}. Classic examples include `gay', originally meaning `happy' or `cheerful', now used to describe a sexual orientation; and  `awful', originally meaning majestic, then used as `horrible' before gaining the opposite meaning -- `wonderful' \cite{hamilton2016diachronic}.

Broadly speaking, a semantic shift is a grass-root process -- a result of the way the speaker community use the language, rather than a shift forced by a regulatory body. That is, an institution (e.g., a corporation) may develop an `anti-virus' (referring to a computer virus, rather than an infectious agent that replicates within a host organism) but speakers may not adopt the new use of the word. 
Organized social groups can press for a change in the ways words should/not be used (e.g., negro, retard, or guys) but these efforts are not institutionalized. 
Standardization enforced by language regulatory bodies tend to fail, as famously illustrated by the evolution of the French negation \cite{jespersen1917negation,dahl1979typology,hock2009language}. 

\vspace{-0.2cm} 
\paragraph{Emoji vs. Written Text} Emoji is unique in the way it combines four elements, each is governed by a different mechanism: (i) a subtle contextual meaning (like many proper words in the language), (ii) a strict ``spelling'' (the Unicode code), (iii) a formal description (e.g., ``face with tears of joy'') decided by a regulatory body (the Unicode Consortium), and (iv) a flexible visual standard decided by any number of a third-party organizations. Together, these elements restrict the ways in which emoji can be creatively manipulated by the users. While users can write `coz' instead of the formal dictionary entry `because', or `on da fon wit ma ma' instead of ``on the phone with my mother'', they cannot be playful with the unicode of a specific smiley from the `emoji dictionary'. Creativity is typically achieved by emoji collocations and by allusions inspired by visual features. The eggplant emoji \includegraphics[height=1em]{figs/emojis/aubergine_1f346_iOS_13_2.png}, for example, is commonly used as a sexual reference to the male genitalia (replacing the \includegraphics[height=1em]{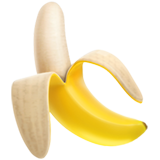}, previously used for the same purpose).  


\section{Related Work}
\label{sec:related}

\paragraph{Emoji Echo-system and Representation} The emoji echo-system, its cultural and historical roots, its evolution and the socio-linguistic function of the use of emoji are surveyed in \cite{evans2017emoji,mcculloch2019because}. Cultural differences in the use of emoji are studies by \cite{chen2018through,guntuku2019studying,li2019exploring}. The ambiguous interpretation of emoji is studied in \cite{miller2016blissfully}, and temporal variations in usage and popularity are studied by \cite{ai2017untangling,barbieri2018exploring}.

Recent works use neural models for learning emoji representation in a way useful for downstream tasks. A skip-gram model was used to analyze the use of emoji on Instagram \cite{dimson2015} and Twitter \cite{barbieri2016does}.  Emoji2vec \cite{eisner2016emoji2vec} demonstrates that learning the embeddings based on the formal description of the unicode yeilds better representation than learning the embeddings using the emoji unicode as a textual token. DeepMoji \cite{felbo2017using} uses millions of tweets for training a multi-layer (Bi-LSTM with attention layers) network in learning embeddings for 64 popular emoji. The quality of the representation is reported across tasks and benchmarks (sentiment analysis, emotion detection, sarcasm detection). A CBOW model was used in a comparative study of the use of emoji on Twitter and Sina-Weibo \cite{guntuku2019studying}.

All of the works cited above assume a static meaning of the emoji, although ambiguity or cultural differences are acknowledged. 
Few works do address changes in emoji usage. Temporal variations in usage and popularity of emoji are addressed by \cite{ai2017untangling,barbieri2018exploring}.
To the best of our knowledge, the only work to address semantic change of emoji from a linguistic perspective is \cite{robertson2021semantic}. No prior work study the semantic change emoji with regards to visual differences and graphical changes across time and platforms.

\vspace{-0.2cm} 
\paragraph{Semantic drift} Semantic change is studied by linguists for decades, see \cite{hock2009language} for a survey. The availability of large textual corpora gave rise to the computational study of semantic change. Early work used changes in word frequency in the span of two centuries, inferring grammatical and semantic shifts \cite{michel2011quantitative} and the changing relations between syntax and meaning \cite{goldberg2013dataset}. The change in gender and racial stereotypes through the \nth{20} century was modeled by \cite{garg2018word}. 

Other works use word embeddings to capture semantic change by measuring the mean-shift \cite{kulkarni2015statistically} or the Spearman correlation between time series \cite{hamilton2016diachronic}. The intersection between the sets of $k$ nearest neighbors of a word $w$ in different time spans is demonstrated to provide stable and interpretable results \cite{gonen2020simple}.

The linguistic factors and processes that facilitate the semantic change include word frequency, polysemy, grammatical category, subjectification, and grammaticalization \cite{algeo1977blends,algeo1980all, brinton2005lexicalization,xu2015computational,dubossarsky2015bottom,hamilton2016cultural,dubossarsky2017outta}, and cultural phenomena, such as the emergence of new technologies \cite{hamilton2016cultural}. 


\section{Data}
\label{sec:data}

\paragraph{Data Collection} Using the Twitter streaming API we collected 1\% (daily) of all English tweets containing emoji between May 2016 and April 2019. In total, our data contained over 209 million tweets posted by 37 million unique users. Table \ref{tab:datastats} presents the total number of tweets collected, as well as the tweets per each of the three main platforms used to tweet. We provide the numbers for each of these platforms since the visual realization of emoji may change between platforms and since the monthly subset per platform may impact the models learnt for each month and platform. The total number of users per platform exceeds the total number of users since many users tweet from more than one platform. For example, 1.9 million users use both an iOS device and the web app; 1.7 million use both an Android device and the web app, and 367K users use all three platforms. 

\begin{table}[h!]
\centering\footnotesize
\begin{tabular}{ c c c } 
 \hline
  & Tweets & Users \\ 
  \hline
  \hline
  ALL & 209M & 37M \\
 \hline
 iOS & 113M & 20M \\
 Android & 56M & 13M \\
 Web & 33M & 8.7M \\
 \hline
\end{tabular}
\caption{Counts of English tweets containing Emoji, collected in the span of 5/2016--4/2019. ALL counts refer to all platforms, not only the most popular three. }
\label{tab:datastats}
\end{table}

The longevity of the data covers multiple changes in the emoji visualizations across platforms, including some significant changes such as the pistol (\includegraphics[height=1em]{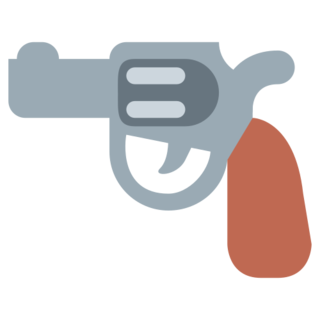} $\mapsto$ \includegraphics[height=1em]{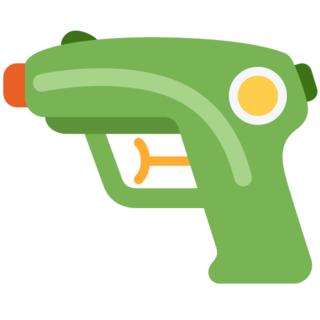}) and the cookie (\includegraphics[height=1em]{figs/emojis/cookie_1f36a_samsung_touchwiz7_1.png} $\mapsto$ \includegraphics[height=1em]{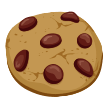}), among many other less pronounced changes, e.g., the ROFL emoji (\includegraphics[height=1em]{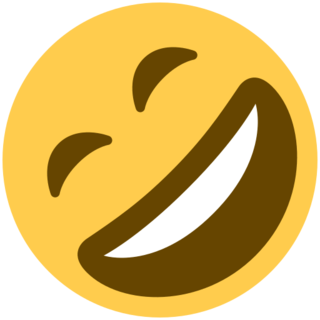} $\mapsto$ \includegraphics[height=1em]{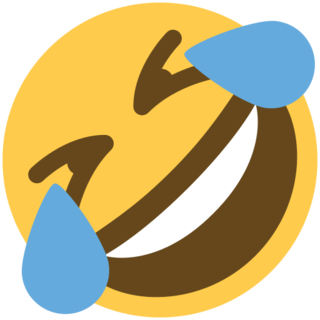}). These changes provide the unique opportunity to study the effect of the visual realization on the semantics of emoji, as well as  drifts unrelated to a forced visual change\footnote{It is important to note that some visual differences more pronounced than others (consider \includegraphics[height=1em]{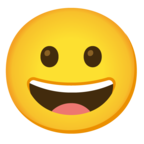},\includegraphics[height=1em]{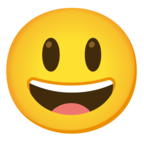},\includegraphics[height=1em]{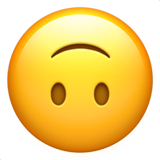},\includegraphics[height=1em]{figs/emojis/rolling-on-the-floor-laughing_1f923_twemoji_2_1.png} ,\includegraphics[height=1em]{figs/emojis/rolling-on-the-floor-laughing_1f923_twemoji_12_1_4.png}), thus may not trigger a significant change in the way they perceived after the change. Quantifying the graphical  change and and its cognitive impact is a challenging task. We note, however, that semantic change could serve as a proxy to the impact of the visual change.\label{fn:visual_change}}.

\vspace{-0.2cm} 
\paragraph{Data Preprocessing} Retweets and tweets lacking text were removed from the corpus. The remaining tweets were lower-cased and punctuation was padded. Additionally, we removed URLs, special characters, and user mentions.  Consecutive emoji were padded with white spaces\footnote{In some special cases a sequence of emoji unicodes with no white spaces but with the zero-width-joiner code \texttt{U+200D} serves as a ``modifier'', e.g., the sequence \texttt{U+1F469U+1F469U+1F466} (codes for \texttt{woman},\texttt{woman},\texttt{boy}) is displayed as \includegraphics[height=1em]{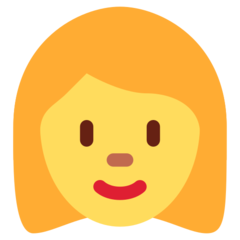}\includegraphics[height=1em]{figs/emojis/woman_1f469.png}\includegraphics[height=1em]{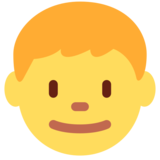}, 
while the sequence \texttt{U+1F469U+200DU+1F469U+200DU+1F466} is displayed as a single emoji -- a family unit of a gay/lesbian couple and a boy \includegraphics[height=1em]{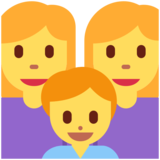}, a modification of the \texttt{U+1F46A} (\texttt{family}) emoji~\includegraphics[height=1em]{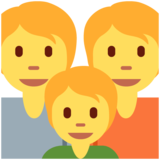}. We keep these sequences unpadded, in order to maintain the modified emoji as a single emoji token. }.


\section{Experimental Setting}
\label{sec:experimental_setting}

\subsection{Learning Embedding Models}
 \paragraph{Embedding Models} Given $n$ data sets ($n=T\times P$; for $T$ time-spans and $P$ platforms), we train an embedding model\footnote{All reported results were obtained using the skip-gram model \cite{mikolov2013distributed}.} $M^i$ ($i=1,...,n$) for each dataset. 
 In total, we trained $36\times 3$ models -- a model for each month and platform in our data.

\vspace{-0.2cm} 
\paragraph{Sanity Checks} Since our models are trained on $n$ relatively small subsets of the data (e.g., an average of less than a million tweets per monthly model for the Android platform, see Table \ref{tab:datastats}), we first verify that the models indeed learn proper embeddings. We do this by applying a set of standard analogy tests\footnote{To the degree analogy tests do learn inference relations, and see \cite{levy2015supervised}.}. The test contained both word analogies (e.g., $\overrightarrow{king}-\overrightarrow{man}+\overrightarrow{woman}=\overrightarrow{queen}$) and emoji analogies (e.g., \includegraphics[height=1em]{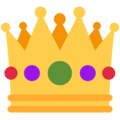} $-$ \includegraphics[height=1em]{figs/emojis/woman_1f469.png} $+$\includegraphics[height=1em]{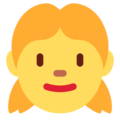} $=$ \includegraphics[height=1em]{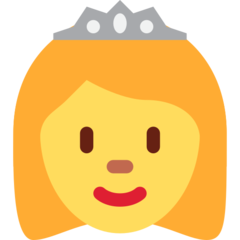}\footnote{There is a crown, prince and princes emoji but no emoji for king or queen, so we use the crown, woman, girl and princess emoji.}).

\subsection{Quantifying Pair-wise Drift}
\label{subsec:pairwise_drift}
  Given an embedding model $M^i$ and a vocabulary $V$ (shared across datasets and derived models), the matrix $D^i$ holds the pair-wise distances between all pairs of words (and emoji) in dataset $i$, so $D^i_{kl}= d(v_k,v_l)$ for a given distance function $d$, e.g., cosine-similarity, and the embeddings of $v_k,v_l$ in $M^i$. 

The baseline shift between two models $M^i$ and $M^j$ is given by the matrix $D^{i\mapsto j}=D^i-D^j$ (all values are converted to their absolute value). The mean-shift and variance of the baseline shift, $\mu^{i \mapsto j}$ and  $\sigma^{i \mapsto j}$, are computed over the values of $D^{i\mapsto j}$.

The binary matrix $\Delta^{i\mapsto j}$ indicates the {\em pair-wise} semantic shift for all pairs of words, where the shift for a pair $\{v_k, v_l\}$ is defined as:   
\vspace{-0.3cm}
\begin{equation*}
\Delta^{i\mapsto j}_{kl} = \begin{cases}
1 &\text{$|D^{i\mapsto j}_{kl} - \mu^{i \mapsto j}| > \beta\sigma^{i \mapsto j}$}\\
0 &\text{Otherwise}
\end{cases}
\label{eq:delta}
\end{equation*}

for some $\beta \geq 2$, establishing the statistical significance of the shift\footnote{We use the Shapiro-Wilk test to verify the normality of the distribution of the values of $\Delta^{i \mapsto j}$.}. 

While $\Delta^{i\mapsto j}_{kl}$ indicates the pair-wise shift, we are interested in identifying which of the words in the pair underwent the shift. In order to do that we simply choose the word that changed with respect to more words. That is, for each $k,l$ for which $\Delta^{i\mapsto j}_{kl}=1$, we decide whether it is $k$ or $l$ that shifted by computing:
\vspace{-0.3cm}
\begin{equation*}
\phi(k,l,\Delta^{i\mapsto j}) = \begin{cases}
k & \Sigma_{p \neq k} \Delta^{i\mapsto j}_{kp}  > \Sigma_{p \neq l} \Delta^{i\mapsto j}_{pl} 
\\
l & \Sigma_{p \neq k} \Delta^{i\mapsto j}_{kp}  < \Sigma_{p \neq l} \Delta^{i\mapsto j}_{pl} 
\end{cases}
\end{equation*}

Naturally, it is possible that both $k$ and $l$ have shifted, although $\phi(k,l, \Delta^{i\mapsto j})=k$. The shift of $l$ will be acknowledged by $\phi(p,l,\Delta^{i\mapsto j})$ returning $l$ for some $p\neq k$, that is, the shift of a word is checked for every occurrence of it in $\Delta^{i\mapsto j}$. 


\section{Results and Discussion}
\label{sec:results}
Using the method described in Section \ref{subsec:pairwise_drift}, we identify dozens of drifting emoji. Table \ref{tab:drifts} presents some illuminating examples, along with the semantic context before and after the change. In the remainder of this section we discuss some of these in more detail. 

\begin{table}[h!]
\centering\footnotesize
\begin{tabular}{ c | p{3cm} | p{2.5cm} } 
 \hline
  Token & Before Change & After Change \\ 
  \hline
  \hline
 rally & politics, geo-locations & college, sports \\
 Florence & tourism & hurricanes, disaster \\
 boomer & terror, gaming & condescension, millennials   \\
 \hline
  \includegraphics[height=1em]{figs/emojis/pistol_1f52b_twemoji_1.png} & anger, death, failure, gun rights & games, gaming, online trends \\
 \includegraphics[height=1em]{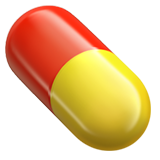} & drug abuse, weapons & medicine\\
 \includegraphics[height=1em]{figs/emojis/cookie_1f36a_iOS_13_2.png} & sweets, desserts & baking, food\\
 \hline
\end{tabular}
\caption{Examples of words and emoji that went through a semantic shift between 2016 and 2019.}
\label{tab:drifts}
\end{table}

\vspace{-0.4cm} 
\paragraph{Expected Drift} Following \cite{kulkarni2015statistically} example's of Twitter drifts ({\em rally}: politics $\mapsto$ gaming; {\em sandy}: beaches $\mapsto$ hurricanes), we explicitly verified the existence of three assumed {\em word} drifts in our data: {\em rally}, {\em Florence}\footnote{Florence is the code of a devastating 2018 hurricane, similar to Sandy in Oct. 2012.}, and the recent {\em [ok] boomer} trend.  Indeed, we identified the drifts, as expected (see Table \ref{tab:drifts}, Top).

\vspace{-0.2cm} 
\paragraph{The Right to Bear Arms:~ \includegraphics[height=1em]{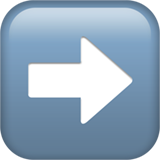}\includegraphics[height=1em]{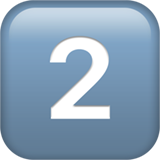}\includegraphics[height=1em]{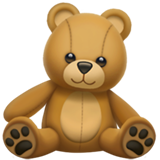}\includegraphics[height=1em]{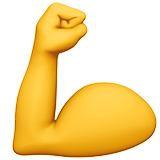}} The pistol emoji presents a unique case, as its visual changed significantly, from a realistic pistol to a toy (\includegraphics[height=1em]{figs/emojis/pistol_1f52b_twemoji_1.png} $\mapsto$ \includegraphics[height=1em]{figs/emojis/pistol_1f52b_twemoji_12_1_4.png}). Indeed, this drift was identified across platforms, in the \emph{respective time of the graphical change} -- Apple changed the emoji in July 2016, while other platforms changed it during 2018. A careful examination of the semantic relations between the pistol and other words and emoji is illuminating. Before the visual change, the tokens most similar to the pistol emoji were {\em fml}\footnote{Acronym for `fuck my life', commonly used to indicate an embarrassment, disappointment or anger.}, {\em \#shotsfired}, \includegraphics[height=1em]{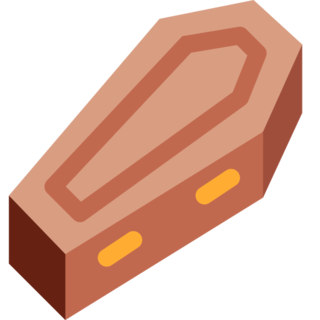} (coffin), \includegraphics[height=1em]{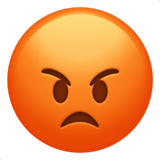} (rage, pouting face), \includegraphics[height=1em]{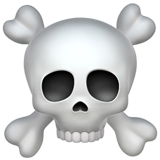} (skull and cross-bones), and {\em 2a} (short for the \nth{2} Amendment, used by gun rights advocates\footnote{The \nth{2} Amendment to the Constitution of the United States stipulating that `` the right of the people to keep and bear Arms, shall not be infringed'' is one of the most divisive issues in American politics.}. Interestingly, we observe a growing semantic gap between the pistol and the former five emoji. We also observe a sharp drop in similarity between the pistol emoji and {\em 2a} in the two months right after the visual change, followed by a slow but stable recovery of the similarity, although it does not reach the semantic similarity of the original emoji. This pattern of recovery deserves a more substantive cultural analysis.   

\vspace{-0.2cm} 
\paragraph{Semantic Drift and Ambiguity} Some of the emoji identified as drifting presented a non stable behaviour -- drifting back and forth through the months without a clear semantic field. Regressing the time series of the change resulted in coefficient that is close to zero. We find that the semantic instability of these emoji correlates with the lack of cohesiveness as measured by the average distance of the emoji from the $k$\footnote{We set $k=50$.} terms (emoji and words) closest to it in the first month, and the average distance between each pair within that set of $k$ terms. We leave further investigation to future work.

\vspace{-0.2cm} 
\paragraph{Seasonality and Global Events} We have identified similar textual drifts as those reported by \cite{kulkarni2015statistically}, although our datasets span different time frames (2011--2013 vs. 2016--2019). These similarities suggest a pattern of cyclic changes, seasonal or related to global events. We observe similar cyclic changes for seasonal emoji like \includegraphics[height=1em]{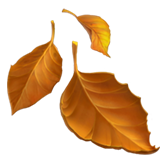} (falling leaves), \includegraphics[height=1em]{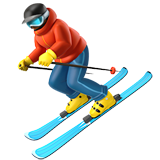} (skier), or emoji related to world singular events like \includegraphics[height=1em]{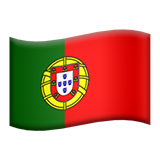} (Portuguese flag) winners of the UEFA EURO 2016. These changes are reversed soon after the season or the event, would it be an election season (rally), a weather catastrophe (Sandy, Florence), or global sport event (Portugal, UEFA). While Kulkarni et al. acknowledge that their Twitter dataset spans a much shorter time, compared to the Google Book Ngram corpus, they do not account for the seasonal or temporal nature of some changes. We believe that these seasonal shifts, of words or emojis, cannot be coupled with the substantive cultural shifts. 
\ignore{Similar cyclic textual drifts were also reported by \cite{gonen2020simple}.}


\section{Limitations and Future Work}
There is a number of limitations to this work, some are inherent to short papers. 

First, reporting on drift in a quantitative way (e.g., ``we find $n$ emoji drifting in our data'' is hard and misleading due to the exploratory nature of the task and the fact that drift is a complex phenomenon -- some drifts are cyclic and some are permanent; some are more pronounced and some are nuanced as the drift occurs on a specific sense of the emoji. We were trying to illustrate the various effects through qualitative examples. This approach is consistent with prior work, for example \cite{garg2018word}.

Second, the literature proposes a number of approaches to measure semantic drift (see Section \ref{sec:related}), each has its pros and cons. In this short paper we propose another measure, inspired by \cite{gonen2020simple}, and apply it on emoji. Future work should address changes in words and study the differences between the different measures. 

Another limitation of the proposed method is a result of the fixed embeddings used. An emoji may have multiple senses, some may drift while others may be stable. Future work should account for the contextual differences in the use of an emoji within the basic time unit.

Finally, we study emoji since they provide a glimpse to the way visual changes affect semantics. However, some emoji went though major changes while other were changed slightly (compare the three visualizations \includegraphics[height=1em]{figs/emojis/cookie_1f36a_iOS_13_2.png}, \includegraphics[height=1em]{figs/emojis/cookie_1f36a_twemoji12_2.png},     \includegraphics[height=1em]{figs/emojis/cookie_1f36a_samsung_touchwiz7_1.png} for the same code \texttt{\footnotesize U+1F36A}). Quantifying the \emph{visual} distance is not a trivial task, although it may be required to get more accurate results (and see Footnote \ref{fn:visual_change}).


\section{Conclusion}
\label{sec:conclusion}
This work presents the first analysis of semantic drift of emoji, accounting for the \emph{graphical variability} of emoji over time and platform. We proposed a novel statistical method for catching drifts and offered a substantive discussion of some of the factors that are unique to emoji use and to the modeling of drift in general. We note that the method we proposed is not specific to emoji and can be applied to any token, as demonstrated in \ref{tab:drifts} (top).

\bibliography{emoji}
\bibliographystyle{acl_natbib}

\end{document}